\documentclass{article}
\usepackage{spconf,amsmath,graphicx}
\usepackage{array} % table use it.
\usepackage{makecell,multirow,diagbox}

\usepackage{cite}
\usepackage{bm}
\usepackage{color}
\usepackage{subfigure}
\DeclareMathOperator*{\argmin}{argmin}

\usepackage[pagebackref=true,breaklinks=true,letterpaper=true,colorlinks,bookmarks=false,citecolor=blue,linkcolor=blue]{hyperref}

\title{AN EFFICIENT DEEP CONVOLUTIONAL LAPLACIAN PYRAMID ARCHITECTURE FOR CS RECONSTRUCTION AT LOW SAMPLING RATIOS}
\name{Wenxue Cui$^1$, Heyao Xu$^1$, Xinwei Gao$^{1,2}$, Shengping Zhang$^1$, Feng Jiang$^1$, Debin Zhao$^1$}
\address{1.Department of Computer Science, Harbin Institute of Technology, Harbin, China  \\
2.Wechat Business Group, Tencent, Shenzhen, China \\
wenxuecui@stu.hit.edu.cn, xuheyao0129@gmail.com, \{s.zhang, fjiang, dbzhao\}@hit.edu.cn   \\
vitogao@tencent.com}
\begin{document}
%\ninept
%
\maketitle
\begin{abstract}
The compressed sensing (CS) has been successfully applied to image compression in the past few years as most image signals are sparse in a certain domain. Several CS reconstruction models have been proposed and obtained superior performance. However, these methods suffer from blocking artifacts or ringing effects at low sampling ratios in most cases. To address this problem, we propose a deep convolutional Laplacian Pyramid Compressed Sensing Network (LapCSNet) for CS, which consists of a sampling sub-network and a reconstruction sub-network. In the sampling sub-network, we utilize a convolutional layer to mimic the sampling operator. In contrast to the fixed sampling matrices used in traditional CS methods, the filters used in our convolutional layer are jointly optimized with the reconstruction sub-network. In the reconstruction sub-network, two branches are designed to reconstruct multi-scale residual images and muti-scale target images progressively using a Laplacian pyramid architecture. The proposed LapCSNet not only integrates multi-scale information to achieve better performance but also reduces computational cost dramatically. Experimental results on benchmark datasets demonstrate that the proposed method is capable of reconstructing more details and sharper edges against the state-of-the-arts methods.
\end{abstract}
\section{Introduction}
\label{sec:intro}

The compressed sensing theory~\cite{candes2006robust, donoho2006compressed} shows that if a signal is sparse in a certain domain $\Psi$, it can be accurately recovered from a small number of random linear measurements less than that of Nyquist sampling theorem. Mathematically, the measurements are obtained by the following linear transformation
\begin{eqnarray}
\label{equ1}
  y = \Phi x + e
\end{eqnarray}
where $x\in R^{N}$ is the signal, $y\in R^{M}$ is known as the measurement vector, $\Phi \in R^{M\times N}$ is the measurement matrix and $e$ denotes noise. If $M\ll N$, reconstructing $x$ from $y$ is generally ill-posed, which is one of the most challenging issues in compressed sensing.

To design an efficient CS reconstruction algorithm, many methods have been proposed, which can be generally divided into two categories: traditional optimization-based methods and recent DNN-based methods.

\begin{figure*}[!t]
\centering
\includegraphics[height=2.2in]{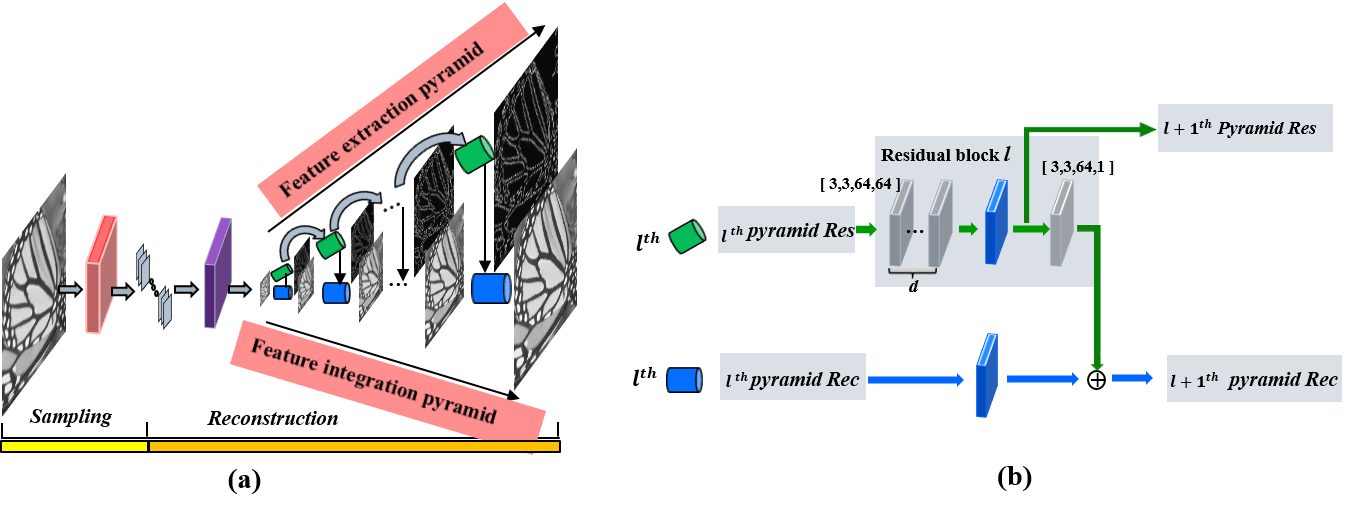}
\caption{(a) is the overview of the proposed LapCSNet and (b) shows the detailed structure in each level. The red box indicates convolutional layer for sampling operator. The sequence of squares indicate measurements. The purple box represents the ``reshape+concat'' layer~\cite{shi2017deep} for initial reconstruction. The gray and blue boxes denote convolutional layers and transposed convolutional layers respectively and the four tuples in the bracket indicate the dimensions of parameters for adjacent convolutional layers.}
\label{figure1}
\end{figure*}

%\textbf{Optimization-based Reconstruction Methods.}
In the optimization-based reconstruction methods, given the linear projections $y$, the original image $x$ can be reconstructed by solving the following convex optimization problem~\cite{candes2006robust, donoho2006compressed}:
\begin{eqnarray}
   \tilde{x} = \mathop{\argmin}_{\mathop{x}}{\frac{1}{2}\|\mathop{\Phi}\mathop{x} - \mathop{y}\|_2^{2}+\lambda\|\mathop{\Psi}\mathop{x}\|_1}
\end{eqnarray}
To solve this convex problem, many algorithms have been proposed~\cite{chen2001atomic, li1tval3, zhang2014group}. However, these algorithms suffer from uncertain reconstruction qualities and high computation cost, which inevitably limit their applications in practice.

%\textbf{DNN-based CS Reconstruction Methods.}
Recently, some DNN-based algorithms have been proposed for image CS reconstruction. In~\cite{mousavi2015deep}, Mousavi et al. propose to utilize a stacked denoising autoencoder (SDA) to reconstruct original images from their measurements. A series of convolutional layers are adopted in~\cite{kulkarni2016reconnet,adler2016deep,shi2017deep} for image reconstruction. Despite their impressive results, massive block artifacts and ringing effects are delivered at low sampling ratios.

To overcome the shortcomings of the aforementioned methods, we propose a Laplacian Pyramid based deep architecture for CS reconstruction. Our network contains two sub-networks: sampling sub-network and reconstruction sub-network. In the sampling sub-network, a convolutional layer with the kernel size of $B\times B$ ($B$ is the size of current block) is utilized to mimic the sampling process. In the reconstruction sub-network, we use two branches to reconstruct multi-scale residual features and multi-scale target images progressively through Laplacian pyramid architectures. Besides, the second branch integrates multi-scale information from the first branch to preserve finer textures. The sampling sub-network and reconstruction sub-network are optimized jointly with the robust Charbonnier loss~\cite{bruhn2005lucas}.

%Our algorithm differs from existing DNN-base methods in the following three aspects:
%\begin{itemize}
%\item A set of cascaded sub-network for laplacian pyramid is used for CS reconstruction. Smaller version compared with target image is produced firstly and then reconstruct target image progressively using upsampling method. This architecture not only integrates multi-scale information sufficiently but also restricts computational complexity greatly even deeper model or higher definition input.
%\item Residual block is developed for residual learning \cite{he2016deep} to speed up training period and improve performance.
%\item We train the network with the robust Charbonnier loss function \cite{bruhn2005lucas} in an end-to-end fashion which promotes visual quality remarkably.
%\end{itemize}
\section{PROPOSED METHOD}
\label{sec:format}

In this section, we describe the methodology of the proposed LapCSNet including the sampling sub-network and the reconstruction sub-network as well as the loss function.

\begin{figure*}[tb]
%\setlength{\abovecaptionskip}{1.0pt}
%\setlength{\belowcaptionskip}{-2.9pt}
%\centering
\begin{minipage}[t]{0.12\textwidth}
\centering
\includegraphics[width=0.86in]{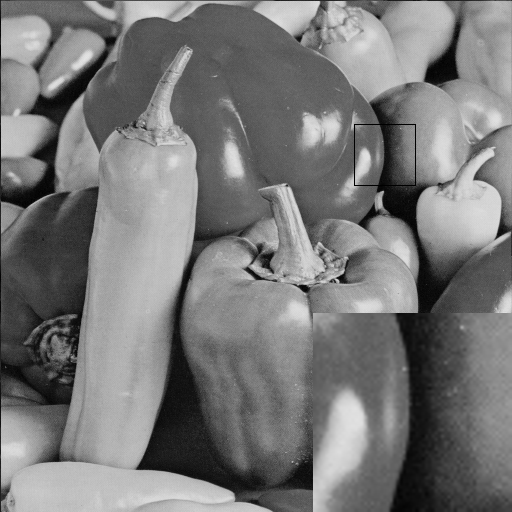}
\begin{scriptsize}
%\begin{spacing}{-0.8}
\centering
\vskip -0.47 cm \begin{tiny}Original$\backslash$PSNR$\backslash$SSIM\end{tiny}
%\end{spacing}
\end{scriptsize}
\end{minipage}
\hfill
\begin{minipage}[t]{0.12\textwidth}
\centering
\includegraphics[width=0.86in]{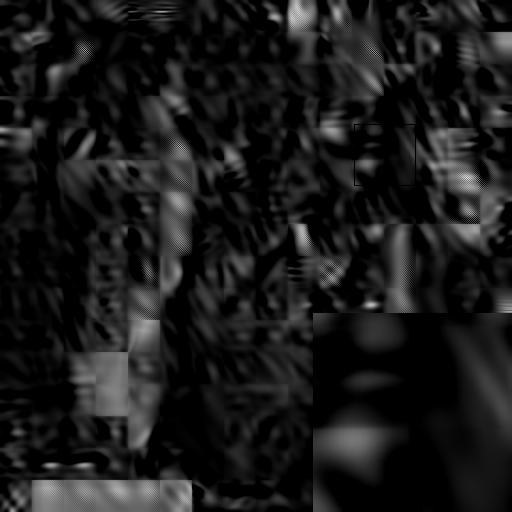}
\begin{scriptsize}
%\begin{spacing}{-0.8}
\centering
\vskip -0.47 cm \begin{tiny}HM$\backslash$12.59$\backslash$0.1586\end{tiny}
%\end{spacing}
\end{scriptsize}
\end{minipage}
\hfill
\begin{minipage}[t]{0.12\textwidth}
\centering
\includegraphics[width=0.86in]{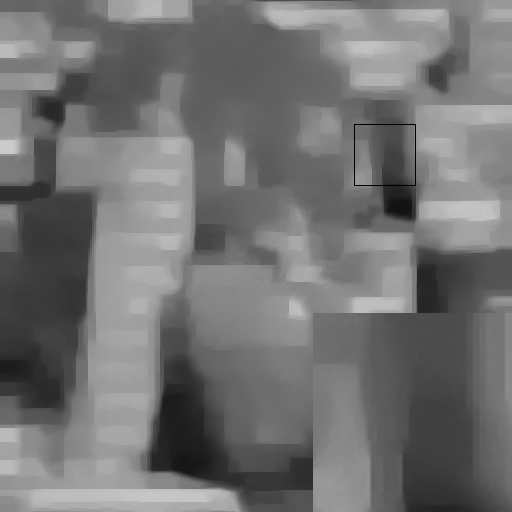}
\begin{scriptsize}
%\begin{spacing}{-0.8}
\centering
\vskip -0.47 cm \begin{tiny}Cos$\backslash$19.13$\backslash$0.5987\end{tiny}
%\end{spacing}
\end{scriptsize}
\end{minipage}
\hfill
\begin{minipage}[t]{0.12\textwidth}
\centering
\includegraphics[width=0.86in]{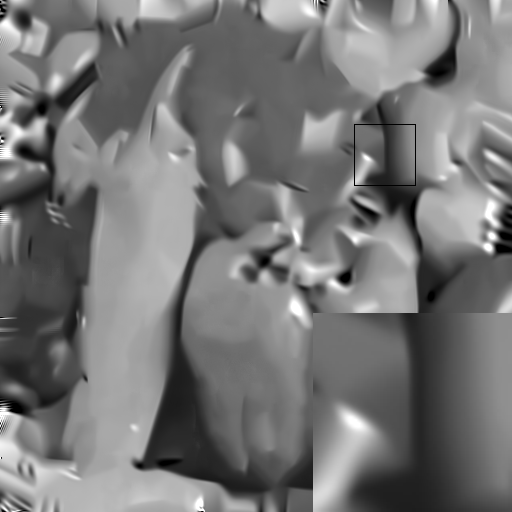}
\begin{scriptsize}
%\begin{spacing}{-0.0}
\centering
\vskip -0.47 cm \begin{tiny}GSR$\backslash$21.01$\backslash$0.6083\end{tiny}
%\end{spacing}
\end{scriptsize}
\end{minipage}
\hfill
\begin{minipage}[t]{0.12\textwidth}
\centering
\includegraphics[width=0.86in]{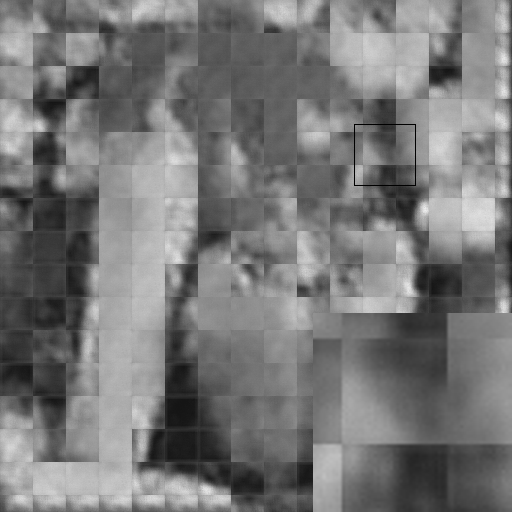}
\begin{scriptsize}
%\begin{spacing}{-0.0}
\centering
\vskip -0.47 cm \begin{tiny}ReconNet$\backslash$19.96$\backslash$0.5726\end{tiny}
%\end{spacing}
\end{scriptsize}
\end{minipage}
\hfill
\begin{minipage}[t]{0.12\textwidth}
\centering
\includegraphics[width=0.86in]{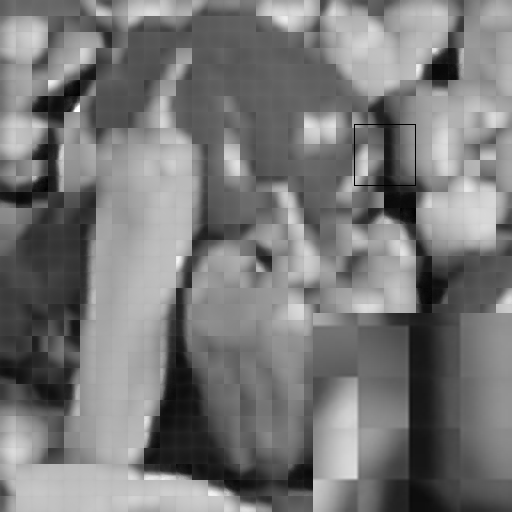}
\begin{scriptsize}
%\begin{spacing}{-0.8}
\centering
\vskip -0.47 cm \begin{tiny}BCSNet$\backslash$23.46$\backslash$0.6508\end{tiny}
%\end{spacing}
\end{scriptsize}
\end{minipage}
\hfill
\begin{minipage}[t]{0.12\textwidth}
\centering
\includegraphics[width=0.86in]{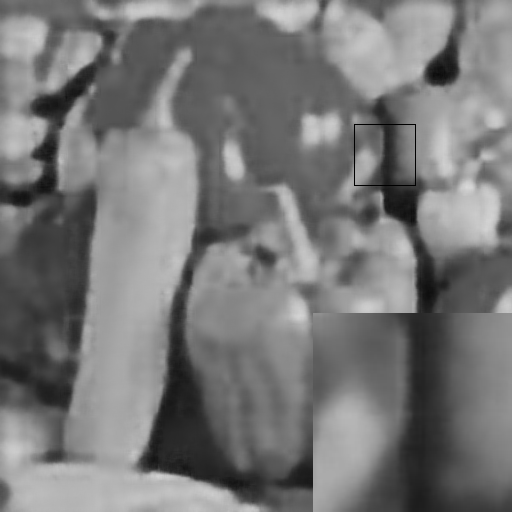}
\begin{scriptsize}
%\begin{spacing}{-0.8}
\centering
\vskip -0.47 cm \begin{tiny}CSNet$\backslash$25.73$\backslash$0.7209\end{tiny}
%\end{spacing}
\end{scriptsize}
\end{minipage}
\hfill
\begin{minipage}[t]{0.12\textwidth}
\centering
\includegraphics[width=0.86in]{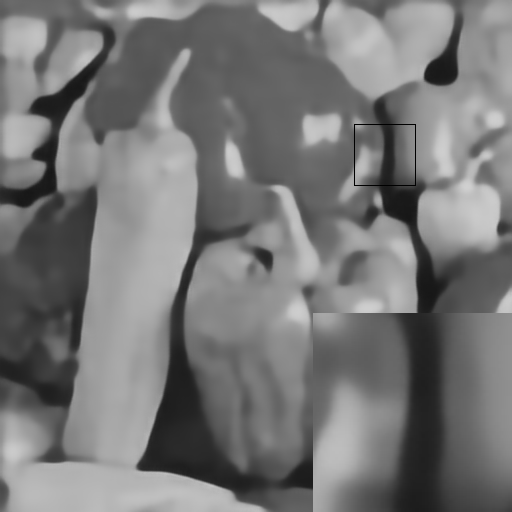}
\begin{scriptsize}
%\begin{spacing}{-0.8}
\centering
\vskip -0.47 cm \begin{tiny}Ours$\backslash$\textbf{26.68}$\backslash$\textbf{0.7588}\end{tiny}
%\end{spacing}
\end{scriptsize}
\label{figure2}
\end{minipage}
\vspace{-0.5em} \caption{Visual quality comparison of image CS recovery on image \emph{Pepper} from Set14 in the case of sampling ratio = 0.01}
\label{figure2}
\end{figure*}

\begin{figure*}[tb]
\setlength{\abovecaptionskip}{1.0pt}
\setlength{\belowcaptionskip}{-2.9pt}
%\centering
\begin{minipage}[t]{0.12\textwidth}
\centering
\includegraphics[width=0.86in]{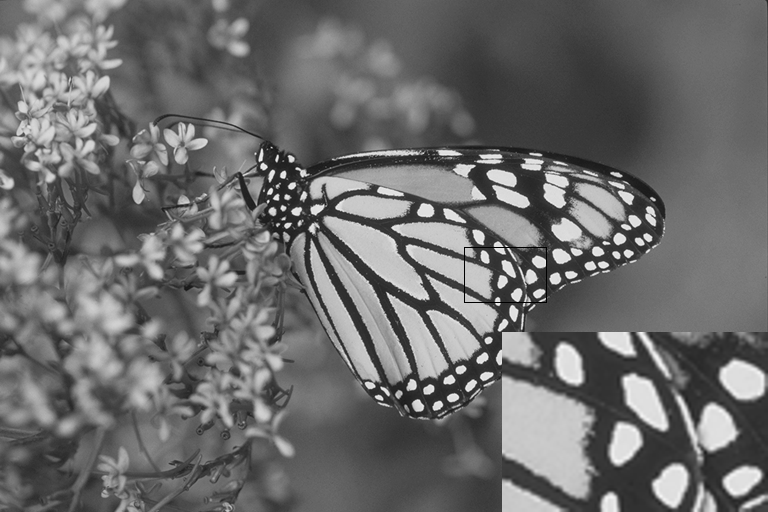}
\begin{scriptsize}
%\begin{spacing}{-0.8}
\centering
\vskip -0.47 cm \begin{tiny}Original$\backslash$PSNR$\backslash$SSIM\end{tiny}
%\end{spacing}
\end{scriptsize}
\end{minipage}
\hfill
\begin{minipage}[t]{0.12\textwidth}
\centering
\includegraphics[width=0.86in]{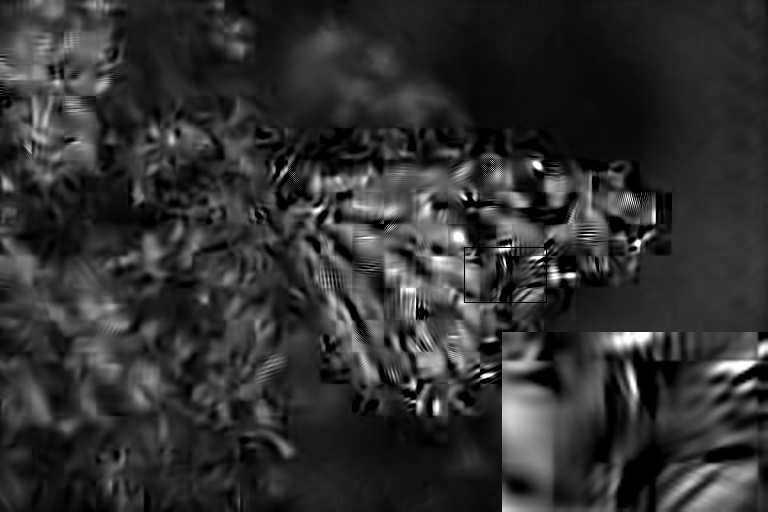}
\begin{scriptsize}
%\begin{spacing}{-0.8}
\centering
\vskip -0.47 cm \begin{tiny}HM$\backslash$18.85$\backslash$0.6909\end{tiny}
%\end{spacing}
\end{scriptsize}
\end{minipage}
\hfill
\begin{minipage}[t]{0.12\textwidth}
\centering
\includegraphics[width=0.86in]{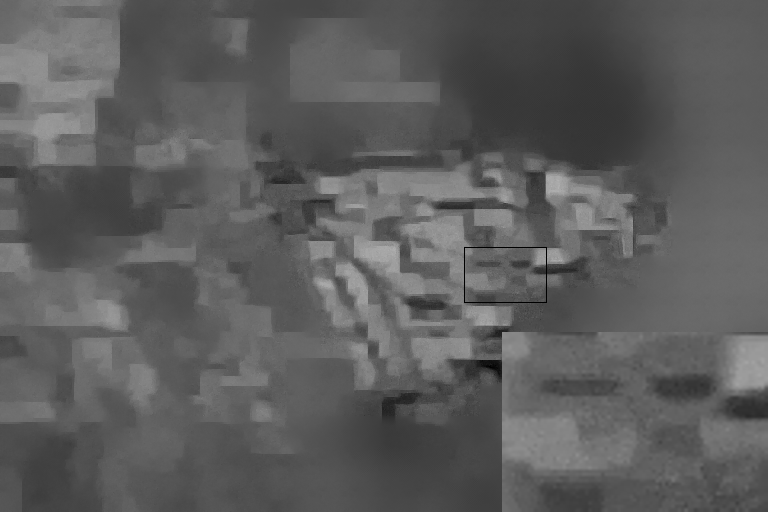}
\begin{scriptsize}
%\begin{spacing}{-0.8}
\centering
\vskip -0.47 cm \begin{tiny}Cos$\backslash$17.66$\backslash$0.6546\end{tiny}
%\end{spacing}
\end{scriptsize}
\end{minipage}
\hfill
\begin{minipage}[t]{0.12\textwidth}
\centering
\includegraphics[width=0.86in]{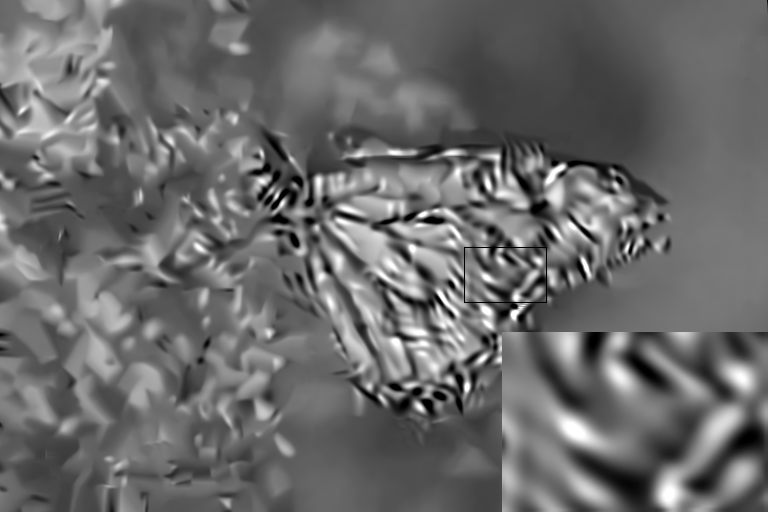}
\begin{scriptsize}
%\begin{spacing}{-0.0}
\centering
\vskip -0.47 cm \begin{tiny}GSR$\backslash$20.74$\backslash$0.6832\end{tiny}
%\end{spacing}
\end{scriptsize}
\end{minipage}
\hfill
\begin{minipage}[t]{0.12\textwidth}
\centering
\includegraphics[width=0.86in]{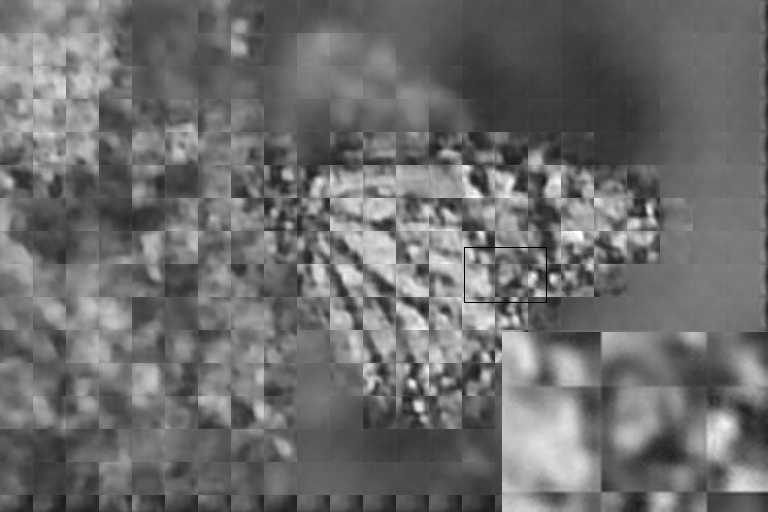}
\begin{scriptsize}
%\begin{spacing}{-0.0}
\centering
\vskip -0.47 cm \begin{tiny}ReconNet$\backslash$20.00$\backslash$0.6706\end{tiny}
%\end{spacing}
\end{scriptsize}
\end{minipage}
\hfill
\begin{minipage}[t]{0.12\textwidth}
\centering
\includegraphics[width=0.86in]{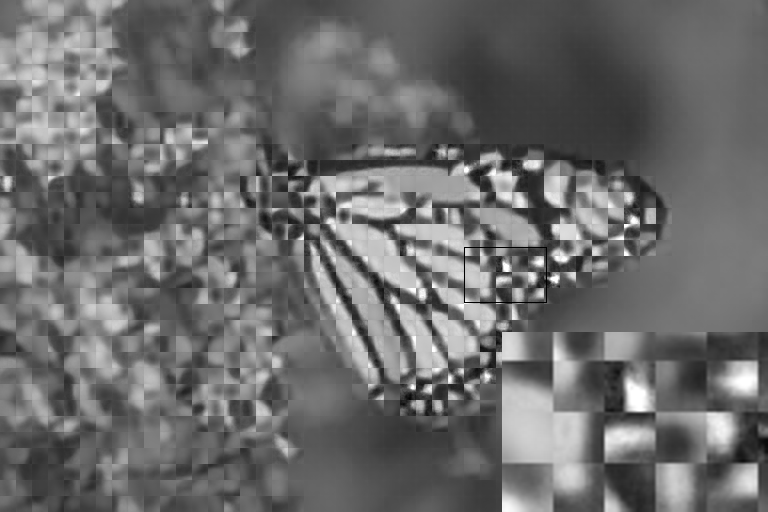}
\begin{scriptsize}
%\begin{spacing}{-0.8}
\centering
\vskip -0.47 cm \begin{tiny}BCSNet$\backslash$22.57$\backslash$0.7531\end{tiny}
%\end{spacing}
\end{scriptsize}
\end{minipage}
\hfill
\begin{minipage}[t]{0.12\textwidth}
\centering
\includegraphics[width=0.86in]{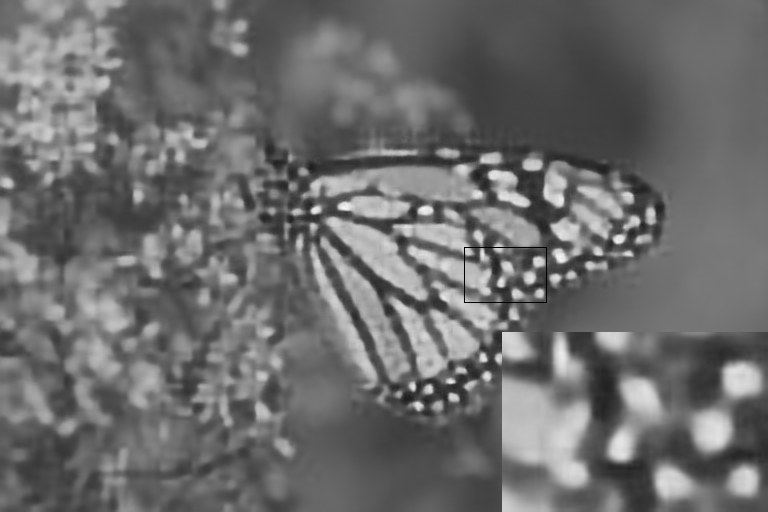}
\begin{scriptsize}
%\begin{spacing}{-0.8}
\centering
\vskip -0.47 cm \begin{tiny}CSNet$\backslash$24.90$\backslash$0.8098\end{tiny}
%\end{spacing}
\end{scriptsize}
\end{minipage}
\hfill
\begin{minipage}[t]{0.12\textwidth}
\centering
\includegraphics[width=0.86in]{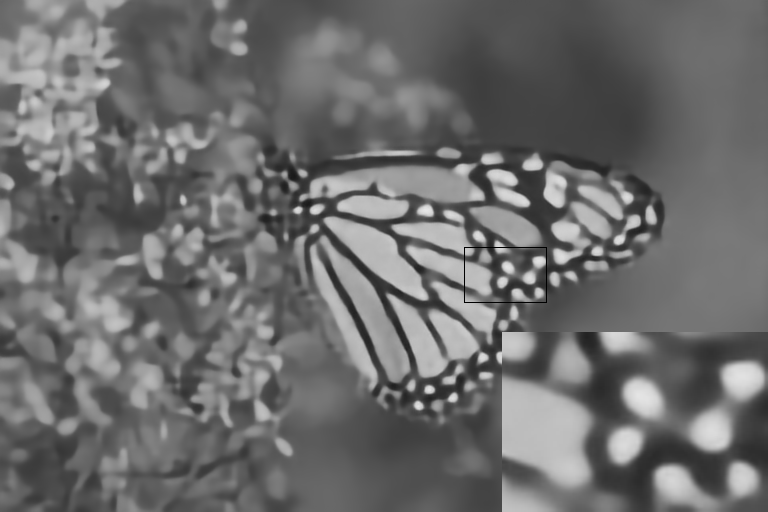}
\begin{scriptsize}
%\begin{spacing}{-0.8}
\centering
\vskip -0.47 cm \begin{tiny}Ours$\backslash$\textbf{25.57}$\backslash$\textbf{0.8351}\end{tiny}
%\end{spacing}
\end{scriptsize}
\label{figure3}
\end{minipage}
\vspace{-0.5em} \caption{Visual quality comparison of image CS recovery on image \emph{Monarch} from Set14 in the case of sampling ratio = 0.02}
\label{figure3}
\end{figure*}

\subsection{Sampling Sub-network}
In traditional block-based compressed sensing (BCS), each row of the sampling matrix $\Phi$ can be considered as a filter. Therefore, the sampling process can be mimicked using a convolutional layer~\cite{adler2016deep,shi2017deep}. In our model, we use a convolutional layer with $B\times B$ filters and set stride as $B$ for non-overlapping blocks. Specifically, given an image with size $w \times h$, there are a total of $L = \lfloor\frac{w}{B}\rfloor\times \lfloor\frac{h}{B}\rfloor$ non-overlapping blocks with size $B \times B$ ($B = 32$). The dimensions of measurements for each block is $n_{B}=\lfloor\frac{M}{N}B^{2}\rfloor$, Therefore, the dimensions of measurements for the current image is $L \times n_{B}$. Traditional sampling matrices are fixed for various reconstruction algorithms. The proposed DNN-based sampling matrices are learned jointly with the reconstruction sub-network from large amounts of data.
\subsection{Reconstruction Sub-network}
For the CS reconstruction, several DNN-based models have been proposed~\cite{kulkarni2016reconnet,adler2016deep}. These methods are implemented for each block, which ignore the relationship between blocks and therefore results in serious blocking artifacts in most cases. To solve this problem, we adopt a ``reshape+concat'' layer~\cite{shi2017deep} to concatenate all blocks to obtain initial reconstruction, which is then refined to obtain superior reconstruction.

\textbf{Initial reconstruction}
Given the compressed measurements, the initial reconstruction block $\tilde{x}_{(j,S)}$ can be obtained by
\begin{eqnarray}
\label{equ2}
  \tilde{x}_{(j,S)} = \tilde{\Phi}_{(B,S)}y_{j}
\end{eqnarray}
where $y_{j}$ is the measurement of the current $j^{th}$ block and $S$ is the scale factor for current block. $\tilde{\Phi}_{(B,S)}$ is a $(\frac{B}{S})^{2}\times n_{B}$ matrix. In the  aforementioned
methods without scaling ($S=1$), is difficult to be accurate calculated. By introducing the pyramid structure, a small block is first reconstructed in our method. Let $\mathcal{Q} = \{2^i | i=1, 2, \ldots, \}$ and $\mathcal{S} = \{s| s\in \mathcal{Q} \quad and \quad \frac{1}{2^s} > \frac{M}{N}\}$. The optimal scale factor $S$ is obtained as
\begin{eqnarray}
\label{qeu4}
  S^{\ast} = Max(\mathcal{S})
\end{eqnarray}
where $Max (\cdot)$ is used to return the maximal element of a set. Therefore, $S>1$  in our model. The convolution output of an image block in the sampling sub-network is a $n_{B}\times 1$ vector, so the size of the convolution filter in the initial reconstruction layer is $1\times 1\times n_{B}$. We use $1\times 1$ stride convolution to reconstruct each block. Since a smaller version of the target block is reconstructed, $(\frac{B}{S})^{2}$ convolution filters of size $1\times 1\times n_{B}$ are used. However, the reconstructed outputs of each block is still a vector. To obtain the initial reconstructed image, a ``reshape+concat'' layer is adopted. This layer first reshapes each $(\frac{B}{S})^{2}$ reconstructed vector into a $\frac{B}{S} \times \frac{B}{S}$ block, then concatenates all the blocks to get the reconstructed image. From Eq. (4), we can get that when $\frac{M}{N}>0.25$, we can not obtain the smaller version of the target image. In this case, we only use this model for CS reconstruction at low sampling ratios.

\textbf{Further reconstruction}
CSNet~\cite{shi2017deep} has only 5 layers for CS reconstruction, which results in poor performance at low sampling ratios. Deep networks and elaborated architectures are essential for accurate reconstruction while increasing computational complexity due to the unenlightened superposition of convolutional layers. Our further reconstruction network takes the output of initial reconstruction as input and progressively predicts residual images at a total of $log_{2}S$ levels where $S$ is the scale factor from Eq.~\ref{qeu4}. Moreover, our reconstruction model has two branches: residual feature extraction~\cite{he2016deep} and feature integration.

$(1)$ Multi-scale residual feature extraction: There is a one-to-one correspondence between reconstruction levels and residual blocks (shown in Fig.~\ref{figure1}). In level $l$, the corresponding residual block consists of $d$ convolutional layers and one transposed convolutional layer~\cite{dumoulin2016guide} to upsample the extracted features by a scale of $2$. The output of each transposed convolutional layer is connected to two different layers: $(i)$ a convolutional layer for reconstructing a residual image for level $l$; $(ii)$ a convolutional layer for extracting features in next level $l+1$. Note that the residual feature at lower levels are shared with higher levels and thus can increase the non-linearity of the network to reconstruct the target image. Moreover, the hierarchical architecture is used for the CS feature extraction, which can preserve more details.

$(2)$ Multi-scale feature integration: In level $l$, the input image is up-sampled by a scale of 2 using a transposed convolutional layer~\cite{dumoulin2016guide}. The upsampled image is then combined with the predicted residual images from the residual feature extraction branch. Then, the output in this level $l$ is fed into the next reconstruction module of level $l$+1. This integration architecture fuses the the multi-scale residual branch and corresponding reconstruction branch efficiently. In addition, most of convolution operation are executed on the smaller version of the target image, which indicates our model is more efficient than the existing models with the same depth.

\begin{table*}[!t]
\centering

    \caption{Quantitative evaluation of state-of-the-arts CS reconstruction algorithms: Average PSNR$\backslash$SSIM$\backslash$time$\backslash$network Layers for sampling ratios 0.01, 0.02, 0.1 on dataset Set5. {\color{red}Red} text indicates the best and {\color{blue}blue} the second best performance}     % NOTE!  caption goes _before_ the table contents !!
    \label{tab:results1}
    \renewcommand\arraystretch{1.10}
    \begin{small}
    \begin{tabular}{>{\hfil}p{55pt}<{\hfil}|>{\hfil}p{98pt}<{\hfil}|>{\hfil}p{98pt}<{\hfil}|
    >{\hfil}p{98pt}<{\hfil}|>{\hfil}p{98pt}<{\hfil}}

    %\begin{tabular}{|c| p{1cm} p{1cm} c| p{1cm} p{1cm} c| p{1cm} p{1cm} c| p{1cm} p{1cm} c|}
    \hline
    %{\bfseries Font} & \multicolumn{3} {c|} {\bfseries Appearance (in Times New Roman or Times} \\
    %\cline{2-4}
    { Alg.} & {  sampling ratio 0.01} & {sampling ratio 0.02} & { sampling ratio 0.1} & {Avg.}      \\
    \hline
    TV~\cite{li1tval3}           & 16.31$\backslash$0.4101$\backslash$22.48$\backslash$--    & 17.94$\backslash$0.4439$\backslash$17.56$\backslash$--	& 18.33$\backslash$0.5921$\backslash$6.93$\backslash$--	&  17.53$\backslash$0.4820$\backslash$15.66$\backslash$--        \\

    %\hline
    MH~\cite{chen2011compressed}             & 12.43$\backslash$0.1999$\backslash$84.32$\backslash$--	&   20.62$\backslash$0.5381$\backslash$78.59$\backslash$--  & 28.57$\backslash$0.8211$\backslash$69.27$\backslash$--	&  20.54$\backslash$0.5197$\backslash$77.39$\backslash$--        \\

    %\hline
    Cos~\cite{zhang2012compressed}            & 17.42$\backslash$0.4122$\backslash$30487.56$\backslash$-- 	&	18.46$\backslash$0.4827$\backslash$27453.49$\backslash$--	& 29.55$\backslash$0.8522$\backslash$6433.25$\backslash$--    &  21.81$\backslash$0.5824$\backslash$21454.65$\backslash$--     \\

    %\hline
    GSR~\cite{zhang2014group}            &  20.81$\backslash$0.5128$\backslash$494.34$\backslash$-- 	&	22.78$\backslash$0.5873$\backslash$532.45$\backslash$--	& 29.99$\backslash$0.8654$\backslash$412.85$\backslash$--	&  24.53$\backslash$0.6552$\backslash$479.88$\backslash$--     \\
    %\hline
    ReconNet~\cite{kulkarni2016reconnet}         & 18.07$\backslash$0.4138$\backslash$0.34$\backslash$7	&	20.05$\backslash$0.4927$\backslash$0.41$\backslash$7	& 24.58$\backslash$0.6762$\backslash$0.37$\backslash$\textbf{\color{blue}7}	&  20.90$\backslash$0.5276$\backslash$0.37$\backslash$7     \\

    %\hline
    BCSNet~\cite{adler2016deep}            & 22.07$\backslash$0.5465$\backslash$\textbf{\color{red}0.01}$\backslash$3	&	23.40$\backslash$0.6168$\backslash$\textbf{\color{red}0.01}$\backslash$3	& 30.01$\backslash$0.8837$\backslash$\textbf{\color{red}0.01}$\backslash$3    &  25.16$\backslash$0.6823$\backslash$\textbf{\color{red}0.01}$\backslash$3  	  \\
    %\hline

    CSNet~\cite{shi2017deep}             & 24.04$\backslash$0.6374$\backslash$\textbf{\color{blue}0.03}$\backslash$6	&	25.87$\backslash$0.7069$\backslash$\textbf{\color{blue}0.02}$\backslash$6	& 32.30$\backslash$0.9015$\backslash$0.04$\backslash$6    &  27.40$\backslash$0.7486$\backslash$\textbf{\color{blue}0.03}$\backslash$6  	  \\
    %\hline
    LapCSNet-2           & $\textbf{\color{blue}24.31}\backslash\textbf{\color{blue}0.6537}\backslash$0.07$\backslash$\textbf{\color{blue}17}    &   $\textbf{\color{blue}26.20}\backslash\textbf{\color{blue}0.7397}\backslash$0.05$\backslash$\textbf{\color{blue}12}  & $\textbf{\color{blue}32.34}\backslash\textbf{\color{blue}0.9023}\backslash$\textbf{\color{blue}0.03}$\backslash$\textbf{\color{blue}7}    &  $\textbf{\color{blue}27.62}\backslash\textbf{\color{blue}0.7652}\backslash$0.05$\backslash$\textbf{\color{blue}12}     \\

    %\hline
    LapCSNet-4           &   $\textbf{\color{red}24.42}\backslash\textbf{\color{red}0.6686}\backslash$0.10$\backslash$\textbf{\color{red}23}  &  $\textbf{\color{red}26.45}\backslash\textbf{\color{red}0.7520}\backslash$0.07$\backslash$\textbf{\color{red}16}   & $\textbf{\color{red}32.44}\backslash\textbf{\color{red}0.9047}\backslash$0.05$\backslash$\textbf{\color{red}9}    &  $\textbf{\color{red}27.77}\backslash\textbf{\color{red}0.7751}\backslash$0.07$\backslash$\textbf{\color{red}16}     \\

    \hline
    \end{tabular}
    \end{small}
\end{table*}

\begin{table*}[!t]
\centering

    \vspace{-0.5em}\caption{Quantitative evaluation of state-of-the-arts CS reconstruction algorithms: Average PSNR$\backslash$SSIM$\backslash$time$\backslash$ network Layers for sampling ratios 0.01, 0.02, 0.1 on dataset Set14. {\color{red}Red} text indicates the best and {\color{blue}blue} the second best performance}      % NOTE!  caption goes _before_ the table contents !!
    \label{tab:results2}
    \renewcommand\arraystretch{1.10}
    \begin{small}
    \begin{tabular}{>{\hfil}p{55pt}<{\hfil}|>{\hfil}p{98pt}<{\hfil}|>{\hfil}p{98pt}<{\hfil}|
    >{\hfil}p{98pt}<{\hfil}|>{\hfil}p{98pt}<{\hfil}}

    %\begin{tabular}{|c| p{1cm} p{1cm} c| p{1cm} p{1cm} c| p{1cm} p{1cm} c| p{1cm} p{1cm} c|}
    \hline
    %{\bfseries Font} & \multicolumn{3} {c|} {\bfseries Appearance (in Times New Roman or Times} \\
    %\cline{2-4}
    { Alg.} & {  sampling ratio 0.01} & {sampling ratio 0.02} & { sampling ratio 0.1} & {Avg.}      \\
    \hline
    TV~\cite{li1tval3}            & 15.17$\backslash$0.3691$\backslash$58.43$\backslash$--    &	17.20$\backslash$0.4069$\backslash$47.36$\backslash$--	& 17.96$\backslash$0.5381$\backslash$17.21$\backslash$--	&  16.78$\backslash$0.4380$\backslash$41.00$\backslash$--        \\

    %\hline
    MH~\cite{chen2011compressed}            & 12.26$\backslash$0.1319$\backslash$95.48$\backslash$--	&   19.20$\backslash$0.4923$\backslash$89.37$\backslash$--  & 26.38$\backslash$0.7282$\backslash$70.19$\backslash$--	&  19.28$\backslash$0.4508$\backslash$85.01$\backslash$--        \\

    %\hline
    Cos~\cite{zhang2012compressed}            & 16.73$\backslash$0.3533$\backslash$23563.48$\backslash$-- 	&	18.35$\backslash$0.4074$\backslash$23042.53$\backslash$--	& 27.20$\backslash$0.7433$\backslash$16596.02$\backslash$--    &  20.76$\backslash$0.5013$\backslash$21067.32$\backslash$--     \\

    %\hline
    GSR~\cite{zhang2014group}            & 19.41$\backslash$0.4583$\backslash$1654.39$\backslash$-- 	&	20.89$\backslash$0.4900$\backslash$1486.10$\backslash$--	& 27.50$\backslash$0.7705$\backslash$948.03$\backslash$--	&  22.60$\backslash$0.5729$\backslash$1362.86$\backslash$--     \\
    %\hline
    ReconNet\cite{kulkarni2016reconnet}         & 18.09$\backslash$0.3907$\backslash$1.04$\backslash$7	&	19.46$\backslash$0.4507$\backslash$1.12$\backslash$7	& 22.91$\backslash$0.5974$\backslash$1.23$\backslash$\textbf{\color{blue}7}	&  20.15$\backslash$0.4796$\backslash$1.13$\backslash$7     \\

    %\hline
    BCSNet~\cite{adler2016deep}            & 20.94$\backslash$0.4910$\backslash$\textbf{\color{red}0.05}$\backslash$3	&	22.00$\backslash$0.5557$\backslash$\textbf{\color{red}0.04}$\backslash$3	& 27.33$\backslash\textbf{\color{red}0.8732}\backslash$\textbf{\color{red}0.05}$\backslash$3    &  23.42$\backslash$0.6400$\backslash$\textbf{\color{red}0.05}$\backslash$3  	  \\
    %\hline

    CSNet~\cite{shi2017deep}            & 22.78$\backslash$0.5574$\backslash$\textbf{\color{blue}0.13}$\backslash$6	&	24.33$\backslash$0.6185$\backslash$\textbf{\color{blue}0.12}$\backslash$6	& 28.91$\backslash$0.8119$\backslash$0.14$\backslash$5    &  25.34$\backslash$0.6626$\backslash$\textbf{\color{blue}0.13}$\backslash$6  	  \\
    %\hline
    LapCSNet-2           & $\textbf{\color{blue}23.03}\backslash\textbf{\color{blue}0.5688}\backslash$0.25$\backslash$\textbf{\color{blue}17}    &   $\textbf{\color{blue}24.55}\backslash\textbf{\color{blue}0.6324}\backslash$0.19$\backslash$\textbf{\color{blue}12}  & $\textbf{\color{blue}28.94}\backslash$0.8124$\backslash$\textbf{\color{blue}0.13}$\backslash$\textbf{\color{blue}7}    &  $\textbf{\color{blue}25.51}\backslash\textbf{\color{blue}0.6712}\backslash$0.19$\backslash$\textbf{\color{blue}12}     \\

    %\hline
    LapCSNet-4           & $\textbf{\color{red}23.16}\backslash\textbf{\color{red}0.5818}\backslash$0.39$\backslash$\textbf{\color{red}23}    &   $\textbf{\color{red}24.76}\backslash\textbf{\color{red}0.6454}\backslash$0.25$\backslash$\textbf{\color{red}16}  & $\textbf{\color{red}29.00}\backslash\textbf{\color{blue}0.8147}\backslash$0.17$\backslash$\textbf{\color{red}9}    &  $\textbf{\color{red}25.64}\backslash\textbf{\color{red}0.6806}\backslash$0.24$\backslash$\textbf{\color{red}16}     \\

    \hline
    \end{tabular}
    \end{small}
\end{table*}

%\section{Implementation and training}
%
%In a DNN model, the loss function and the setting of some hyper parameters is very important for training. For the reproduction of our method, the training details are described in this section.

\subsection{Loss function}
Let $x$ be the input image, $\theta_{S}$ is the parameter of the sampling sub-network and $\theta_{R}$  the parameter of reconstruction sub-network. Two mapping function $f_{S}$ and $f_{R}$ are desired to produce accurate measurements and reconstructed image $\hat{y}=f(x;\theta_{S},\theta_{R})$. We denote the residual image in level $l$ by $r_{l}$, the up-scaled image by $x_{l}$ and the corresponding target image by $\hat{y}_{l}$. The desired output images in level $l$ is modeled by $\hat{y}_{l}=x_{l}+r_{l}$. We use the bicubic downsampling method to resize the ground truth image $y$ to $y_{l}$ at each level. We propose to use the robust Charbonnier penalty function for each level. The overall loss function is defined as:
\begin{eqnarray}
\label{equ2}
\ell(\hat{y},y;\theta_{S},\theta_{R})=\frac{1}{N}\sum_{i=1}^N\sum_{l=1}^L\rho(\hat{y_{l}}^{(i)}-y_{l}^{(i)})
\end{eqnarray}
where $\hat{y}_{l}^{(i)}=x_{l}^{(i)}+r_{l}^{(i)}$ and $\rho(x)=\sqrt{x^{2}+\varepsilon^{2}}$. $N$ is the number of training samples, and $L$ is the number of levels in our pyramid.  $\varepsilon$ is empirically set to $1e-3$.

\section{EXPERIMENTAL RESULTS AND ANALYSIS}
%We first discuss the implementation and training details of proposed network. We then compare our LapCSNet with state-of-the-arts algorithms on two benchmark datasets.
\subsection{Implementation and training details}
In the reconstruction sub-network, each convolutional layer consists of 64 kernels with size of $3\times3$. We initialize the convolutional filters using the same method~\cite{he2015delving}. The size of the transposed convolutional filter is $4\times4$. A ReLU layer with a negative slope of 0.2 is subsequent for all convolutional and transposed convolutional layers. We pad zero values around the boundaries before applying convolution to keep the size of all feature maps the same as the input of each level.

We use the training set (200 images) and testing set (200 images) of the BSDS500 database~\cite{arbelaez2011contour} for training, and the validation set (100 images) of BSDS500 for validation. We set the patch size as $128\times128$, and batch size as 64. We augment the training data in three ways: $(i)$ Randomly scale between $[0.75, 1.2]$. $(ii)$ Rotate the images by $90^{\circ}$, $180^{\circ}$, and $270^{\circ}$. $(iii)$ Flip the images horizontally or vertically with a probability of 0.5. We train our model with the Matlab toolbox MatConvNet~\cite{vedaldi2015matconvnet} on a Titan X GPU. The momentum parameter is set as 0.9 and weight decay as $1e-4$. The learning rate is initialized to $1e-6$ for all layers and decreased by a factor of 2 for every 50 epochs. We train our model for 200 epochs and each epoch iterates 1000 times.

%\section{EXPERIMENTAL RESULTS AND ANALYSIS}
\subsection{Comparison with the state-of-the-arts}
\label{sec:typestyle}
We compare our algorithm with seven representative methods, i.e., total variation (TV) method~\cite{li1tval3}, multi-hypothesis (MH) method~\cite{chen2011compressed}, collaborative sparsity (Cos) method~\cite{zhang2012compressed}, group sparse representation (GSR) method~\cite{zhang2014group}, ReconNet~\cite{kulkarni2016reconnet}, BCSNet~\cite{adler2016deep} and CSNet ~\cite{shi2017deep}. In these algorithms, the first four belong to traditional optimization-based methods, while the last three are recent network-based methods. The PSNR and SSIM reconstruction performances at three different sampling ratios: 0.01, 0.02 and 0.1 for the datasets Set5 and Set14 are summarized in Table~\ref{tab:results1} and Table~\ref{tab:results2}, respectively. The ``LapCSNet-2'' denotes $d=2$ and ``LapCSNet-4'' $d=4$. It can be seen from the tables that about 0.2-0.5dB PSNR improvement is obtained on both test datasets Set5 and Set14 when the sampling ratio is very low. Obviously, the proposed method outperform the existing algorithms in low-ratio by a large margin, which fully demonstrates the effectiveness of our model. Moreover, for the ratio=0.01, our model is about 4 times deeper than CSNet, while the running time is just about 3 times longer than that of the CSNet, which demonstrate the Laplacian pyramid is a high-efficiency design for CS reconstruction. The visual comparisons in the case of ratio=0.01 and ratio=0.02 in Fig.~\ref{figure2} and Fig.~\ref{figure3} show that the proposed LapCSNet is able to reconstruct more details and sharper edges without obvious blocking artifacts.

\section{CONCLUSION}
\label{sec:conclusion}
In this work, we propose a deep convolutional network (LapCSNet) within a Laplacian pyramid framework for fast and accurate CS reconstruction. Our model consists of two sub-networks namely sampling sub-network and reconstruction sub-network. In the sampling sub-network, the sampling filters are learned jointly with the reconstruction sub-network. In the reconstruction sub-network, we divide our model into two branches to extract the residual features and integrate target images using laplacian pyramid architectures, respectively. In other words, the reconstruction sub-network progressively predicts high-frequency residuals and integrate multi-scale information in a coarse-to-fine manner. The sampling sub-network and reconstruction sub-network are optimized jointly using a robust Charbonnier loss function. Experimental results show that the proposed LapCSNet is capable of reconstructing more details and sharper edges against several state-of-the-arts algorithms.

\section{ACKNOWLEDGEMENTS}
This work is partially funded by the Major State Basic Research Development Program of China (973 Program 2015CB351804) and the National Natural Science Foundation of China under Grant No. 61572155 and 61672188.

\bibliographystyle{IEEE}
\bibliography{strings}

\end{document}